\documentclass[10pt,twocolumn,letterpaper]{article}

\usepackage[pagenumbers]{cvpr} 

\usepackage[table,xcdraw]{xcolor}
\usepackage{graphicx}
\usepackage{amsmath}
\usepackage{amssymb}
\usepackage{booktabs}
\usepackage{pifont}
\usepackage{multirow}
\usepackage{bm}

\usepackage{algorithm}
\usepackage{algpseudocode}
\usepackage{listings}

\definecolor{codegreen}{rgb}{0,0.6,0}
\definecolor{codegray}{rgb}{0.5,0.5,0.5}
\definecolor{codepurple}{rgb}{0.58,0,0.82}
\definecolor{backcolour}{rgb}{1.0,1.0,1.0}

\lstdefinestyle{code_style}{
  backgroundcolor=\color{backcolour}, commentstyle=\color{codegreen},
  keywordstyle=\color{magenta},
  numberstyle=\tiny\color{codegray},
  stringstyle=\color{codepurple},
  basicstyle=\ttfamily\footnotesize,
  breakatwhitespace=false,         
  breaklines=true,                 
  captionpos=b,                    
  keepspaces=true,                 
  numbers=none,                    
  numbersep=5pt,                  
  showspaces=false,                
  showstringspaces=false,
  showtabs=false,                  
  tabsize=2
}
\usepackage[pagebackref,breaklinks,colorlinks]{hyperref}

\usepackage[capitalize]{cleveref}
\crefname{section}{Sec.}{Secs.}
\Crefname{section}{Section}{Sections}
\Crefname{table}{Table}{Tables}
\crefname{table}{Tab.}{Tabs.}

\def\cvprPaperID{*****} 
\def\confName{CVPR}
\def\confYear{2023}

\begin{document}

\title{Changer: Feature Interaction is What You Need for Change Detection}

\author{Sheng Fang, Kaiyu Li, Zhe Li*\\
Shandong University of Science and Technology, China\\
{\tt\small fangs99@126.com, \{likyoo, lizhe\}@sdust.edu.cn}
\\
\url{https://github.com/likyoo/open-cd}
}
\maketitle

\begin{abstract}
Change detection is an important tool for long-term earth observation missions. It takes bi-temporal images as input and predicts ``where'' the change has occurred. Different from other dense prediction tasks, a meaningful consideration for change detection is the interaction between bi-temporal features. With this motivation, in this paper we propose a novel general change detection architecture, MetaChanger, which includes a series of alternative interaction layers in the feature extractor. To verify the effectiveness of MetaChanger, we propose two derived models, ChangerAD and ChangerEx with simple interaction strategies: 
Aggregation-Distribution (AD) and ``exchange''. AD is abstracted from some complex interaction methods, and ``exchange'' is a completely parameter\&computation-free operation by exchanging bi-temporal features. In addition, for better alignment of bi-temporal features, we propose a flow dual-alignment fusion (FDAF) module which allows interactive alignment and feature fusion. Crucially, we observe Changer series models achieve competitive performance on different scale change detection datasets. Further, our proposed ChangerAD and ChangerEx could serve as a starting baseline for future MetaChanger design.
\end{abstract}

\section{Introduction}
\label{sec:intro}
Change detection is one of the most widely used fundamental technologies in earth vision. Compared to remote sensing (RS) image segmentation, change detection has two advantages in application.
1) In long-term earth observation, rather than predicting all pixels of the whole image, we often only need to focus on the land-cover category in the changed area.
2) In semi-automated applications, change detection is more tolerant of some misdetection.

Specifically, change detection is a pixel-to-pixel task, which takes bi-temporal images as input and predicts ``where'' the change has occurred. Driven by large amounts of RS data \cite{toker2022dynamicearthnet, shen2021s2looking, verma2021qfabric}, change detection models based on ConvNet or Vision Transformer achieve more competitive performance in many complex scenarios. Recently, most Deep Learning (DL)-based change detection methods have been designed in close relation to segmentation models, and focus on some common problems, for instance, the misdetection caused by edges, small targets, and various scales.

The goal of semantic segmentation is to determine a network to fit the target $Y$ as much
as possible, which can be described by minimizing the empirical loss as: $\min \limits_\theta L(F_\theta(X), Y)$. Different from this, change detection takes two inputs, i.e. the bi-temporal images $X_1$ and $X_2$, and it can be described as $\min \limits_\theta L(F_\theta(X_1, X_2), Y)$.

Therefore, a worthwhile consideration is whether the correlation between $X_1$ and $X_2$ should be explored, and if so, how to implement it. For the first issue, there are many studies illustrating the important effects of interaction between homo/hetero-geneous features \cite{wang2020deep, zhang2022lightweight}. And specifically for bi-temporal images, there are style differences between different temporal image domains, which are due to climate change, pre-processing corrections, \etc The domain differences affect the target of interest (\eg building) and the background in different degrees, which, together with the use of siamese network, makes the understanding of the ``change of interest'' ambiguous for the model. 



\begin{figure}[t]
  \centering
   \includegraphics[width=1.0\linewidth]{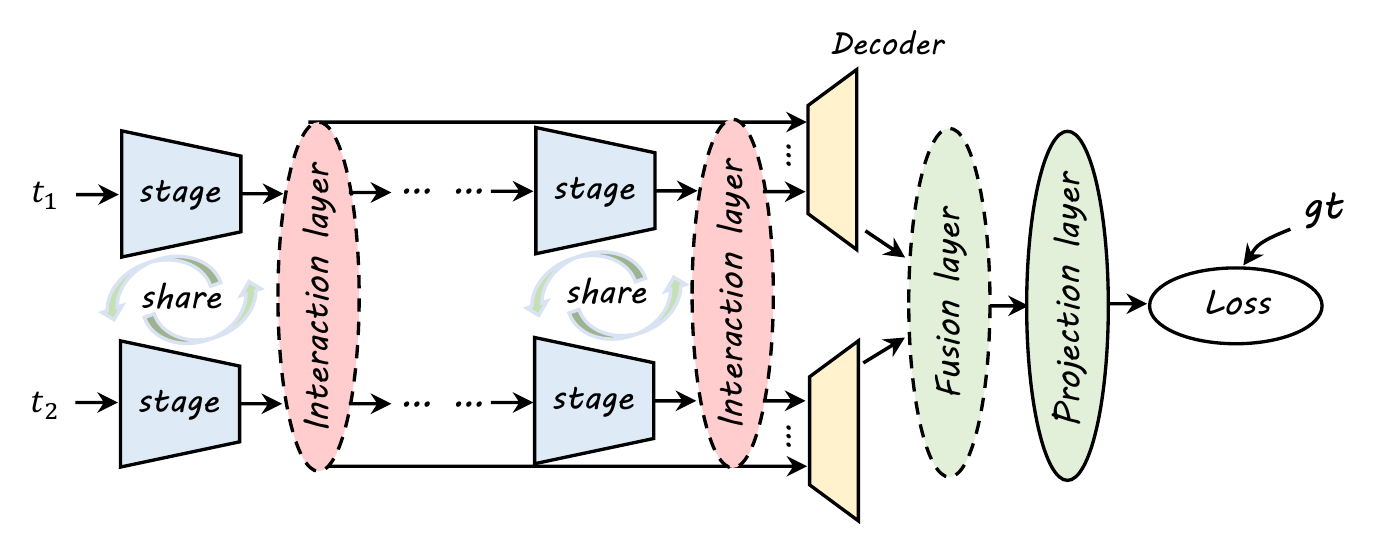}

   \caption{\textbf{MetaChanger for change detection.} MetaChanger is presented as a general architecture with alternative interaction layer (\eg AD and feature ``exchange'' in this paper) and fusion layer (\eg FDAF in this paper). Here, we frame the focus of this paper by dotted lines.``share'' denotes weight sharing.}
   \label{fig:MetaChanger}
\end{figure}

For the second issue, in this paper, firstly, we propose a general change detection architecture, MetaChanger, which aims to emphasize the effect of feature interactions during feature extraction in change detection, as shown in \cref{fig:MetaChanger}. Specifically, there are series of alternative interaction layers and fusion layers designed in MetaChanger.


Then, to verify MetaChanger, we try two simple interaction strategies: aggregation-distribution (AD) and ``exchange''. Specifically, the AD interaction is abstracted from some co-/cross-attention mechanisms, coming from tasks related to multi-modality, tracking, \etc 
And the ``exchange'' interaction is a completely parameter\&computation-free operation, which is achieved by exchanging bi-temporal feature maps in the spatial or channel dimension, with the exchanged features being mixed as they pass through subsequent convolution or token mixer. Astonishingly, these derived model, termed ChangerAD and ChangerEx, achieve extremely competitive performance, and even consistently outperforms other well-tuned change detection models. Moreover, we make further exploration of ChangerEx in view of its excellent performance. The results in multiple datasets show that MetaChanger, even with naïve interaction layers, can still deliver promising performance.

In addition to the interaction during feature extraction, we propose the Flow Dual-Alignment Fusion (FDAF) module for the interactive fusion of dual-branch features to overcome the problem of side-looking and mis-alignment in multi-temporal RS images. All the interaction and fusion components abstracted in MetaChanger are not limited to these specific types. We hope our findings inspire more future research dedicated to improving MetaChanger.

Our main contributions can be summarized as follows.
(1) We propose MetaChanger as a general change detection framework that focuses on a series of alternative interaction layers and can serve as a strong baseline. (2) We propose two embarrassingly simple interaction strategies, AD and feature ``exchange'', and extensive experiments demonstrate that they can greatly improve MetaChanger's performance, especially ``exchange'', even when embedded in complex networks or applied to challenging datasets. (3) We propose an interactive fusion module, called FDAF, which alleviates mis-alignment problem in bi-temporal images.

\begin{figure*}[t]
  \centering
   \includegraphics[width=1.0\linewidth]{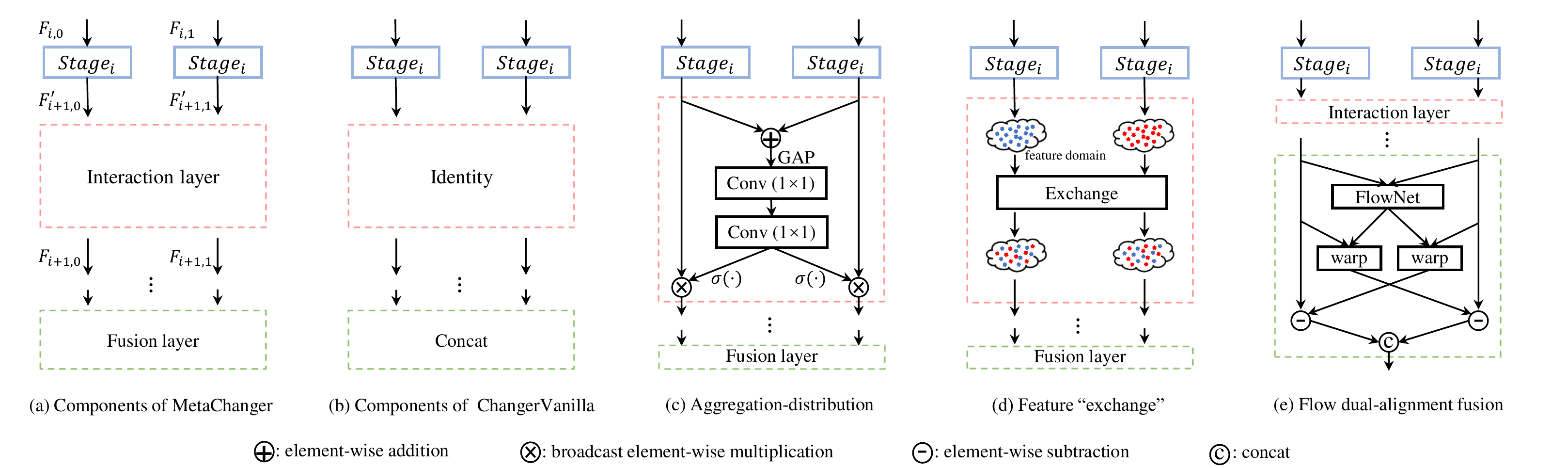}
   \caption{\textbf{Component details of Changer models.} $\sigma(\cdot)$ refers to Sigmoid function and $GAP$ refers to global average pooling. The \bm{$FlowNet$} is consisted of \textbf{DWConv+IN+GELU+PWConv}. (a)$\sim$(e) correspond to Section \ref{section:MetaChanger}$\sim$\ref{section:FDAF}}
   \label{fig:Components}
\end{figure*}

\section{Related work}
\subsection{Binary Change Detection}
We roughly divide DL-based change detection methods into two types, according to how the change map is acquired: metric-based and classification-based. In general, the metric-based method transforms the two input images into a feature space whose feature representation becomes more consistent \cite{shi2021deeply, manas2021seasonal}. In this feature space, the ultimate change map is obtained by the threshold algorithm \cite{liu2016deep}. To achieve this goal, metric learning-related losses, such as contrast loss \cite{hadsell2006dimensionality} are leveraged to pull unchanged pairs together and push changed pairs apart \cite{zhan2017change, chen2020spatial}.

The classification-based method, however, takes change detection as a dense classification task directly \cite{zheng2021change, fang2021snunet, zheng2022changemask, chen2022fccdn}. The bi-temporal features are fused at a certain stage and the ultimate change map is generated by a classifier at the top of the network. Usually, a simple cross-entropy loss is sufficient to optimise the model stably. \cite{daudt2018fully} proposes three typical change detection models: FC-EF, FC-Siam-Conc and FC-Siam-Diff. Among them, FC-EF uses the early fusion strategy and the latter two use the medium fusion strategy with different fusion policies. In addition to CNN models, some transformer-based models achieve competitive performance in change detection. BiT \cite{chen2021efficient} builds an efficient change detection model with very limited parameters by mixing CNN and transformer. ChangeFormer \cite{bandara2022transformerbased} is a pure transformer model, which is a siamese variant of SegFormer \cite{xie2021segformer}, and is fine-tuned in network depth.

Different from the above methods, our proposed MetaChanger, shown in \cref{fig:MetaChanger}, abstracts and simplifies the processes for change detection and especially focuses on the feature interaction between bi-temporal images.

\subsection{Feature Fusion}

Feature fusion is a fundamental process in many DL tasks, such as in classification \cite{sun2019deep, yu2021lite}, segmentation \cite{kirillov2019panoptic, ronneberger2015u}, multimodal tasks \cite{wang2016modality}, \etc They include the fusion of multiple levels, multiple scales, heterogeneous features, \etc

For the specific operation of feature fusion, some simple parameter-free operations such as concat, weighted sum or bilinear pooling \cite{lin2015bilinear} can build stable baseline performance. Additionally, several attention-based methods make feature fusion more flexible and learnable \cite{li2020gated, dai2021attentional, feng2021encoder, bandara2022hypertransformer, zhou2022canet}. Moreover, alignment-based fusion methods focus on feature alignment, and they typically use flow field or deformable convolution \cite{dai2017deformable, zhu2019deformable} to align features of different levels in the spatial dimension \cite{li2020semantic, huang2021alignseg, huang2021fapn, wang2019edvr, tian2020tdan}.

\subsection{Feature Interaction}
In some studies, feature interaction is included in feature fusion. However, here, we define feature interaction in change detection as the correlation or communication of homo/hetero-geneous features during feature extraction before fusion. Co-attention mechanism \cite{lu2016hierarchical} is frequently used in feature interaction, and similar formats have been applied with great success in many research fields like multimodal related tasks (\eg VQA \cite{nguyen2018improved}, RGB-D segmentation \cite{chen2020bi}), Registration \cite{wu2021feature}, Matching \cite{wei2020multi} and network architecture design \cite{li2019selective}. This format of interaction aggregates several features and then distributes them respectively as attention maps. Hence, we abstract this format as AD interaction.

In addition, there are some other effective methods for interaction. For example, \cite{yu2020deformable} uses channel-wise cross-attention to learn mutual information from dual branches for object tracking. \cite{wu2021feature} generates the spatial affinity matrix between source and target point clouds. MixFormer \cite{chen2022mixformer} makes bi-directional interaction across self-attention and DWConv, providing complementary cues in the channel and spatial dimensions. Different from these complex methods, Our proposed feature ``exchange'' method is extremely simple and does not introduce any extra computation.

More relevant to feature “exchange” is CEN \cite{wang2020deep}, which use channel exchange for interaction, where channels corresponding to the smaller BN scaling factors will be replaced by channels from other modalities. Yet, unlike CEN which is applied to multimodal problem, we focus on identifying positions where semantic differences exist for change detection. Hence, strict semantic maintenance and semantic correspondence must be kept between the both domains.


\begin{figure*}[t]
  \centering
   \includegraphics[width=1.0\linewidth]{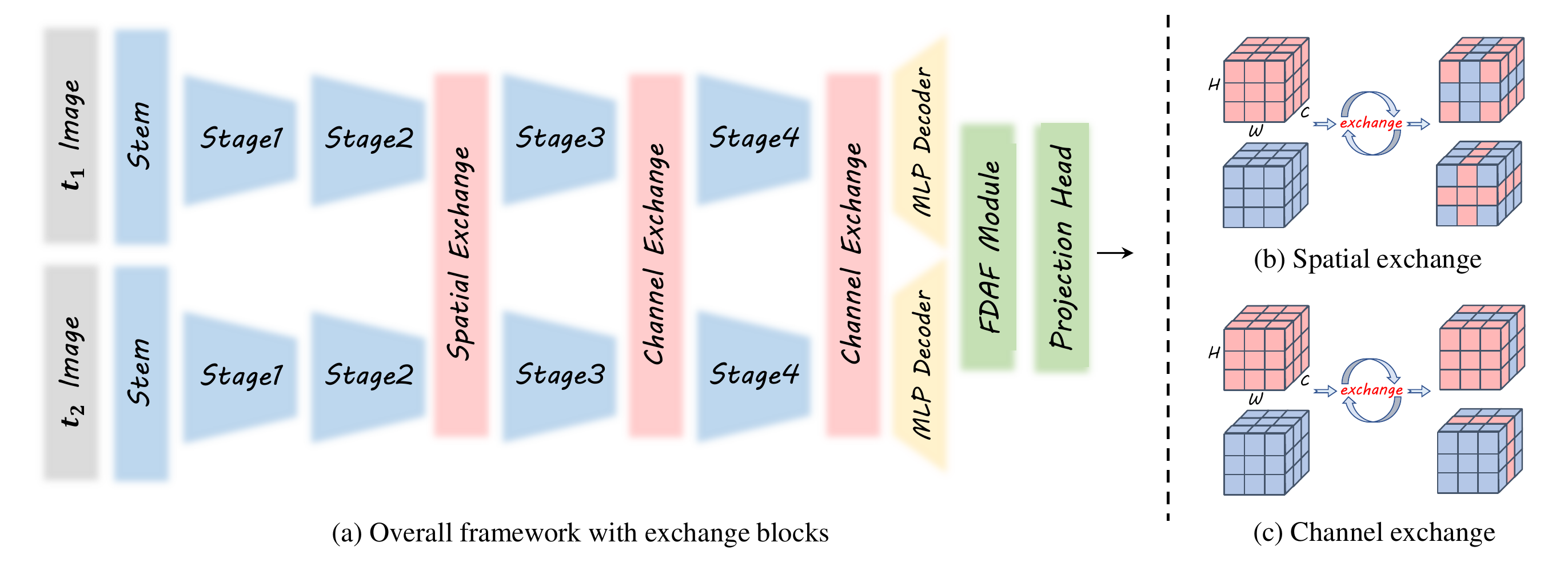}
   \caption{\textbf{ChangerEx model.} (a) ChangerEx adopts spatial exchange in the second stage and channel exchange in the last two stage. FDAF is used to fuse features. (b) The diagram of spatial exchange. (c) The diagram of channel exchange.}
   \label{fig:ChangerEx}
\end{figure*}


\section{Method}
\subsection{MetaChanger}
\label{section:MetaChanger}

MetaChanger is not divorced from classification-/metric-based models, but is a purposeful framework for exploring interaction strategies. In MetaChanger, we use CE loss like the classification-based method to avoid the extra hyper-parameters in the metric-based method.  In addition, to more directly demonstrate the effect of feature interactions, we use an entire siamese encoder-decoder, like the metric-based method. MetaChanger can be indicated as:
\begin{equation}
  Y = H(D(E_{InterAct}(X_1)), D(E_{InterAct}(X_2)))
  \label{eq:3}
\end{equation}

where $X_1$ and $X_2$ indicate the bi-temporal images, $Y$ indicate the ultimate predicted change map; $E_{InterAct}(\cdot)$ and $D(\cdot)$ denote the interactive encoder and decoder network; $H(\cdot)$ denotes the fusion and projection head.
To ensure the generality of MetaChanger, the encoder can be any hierarchical ConvNet or Transformer. Let $F_{i,j}$ denote the output of a hierarchy where $i$ indexes the hierarchy along the encoder and $j$ indexes the temporal dimension ($i\in\{0,1,2,3\}$ typically, and $j\in\{0,1\}$ for change detection). A hierarchy of MetaChanger consists of two main steps. First, multi-level features $F_{i,j}$ go through a network stage with sharing weights and features $F_{i+1,j}^{\prime}$ are generated. Then, $F_{i+1,j}^{\prime},\forall{j}$ feed into the interaction layer to get the correlated $F_{i+1,j}$. The hierarchy $i$ can be formulated as:
\begin{equation}
\begin{aligned}
  & F^{\prime}_{i+1,j} = Stage_i(F_{i,j}), \forall{j} \\
  & F_{i+1,j} = InterAct(F^{\prime}_{i+1,0}, F^{\prime}_{i+1,1}), \forall{j}
  \label{eq:4}
\end{aligned}
\end{equation}
where $InterAct(\cdot)$ refers to the interaction layer. We denote $F^{\prime}_{i,j}$ as $F^{\prime}_{i+1,0}$ and $F^{\prime}_{i+1,1}$ for clearer description in \cref{eq:4}. And to ensure MetaChanger is efficient in practice, we use a light-weight MLP decoder, like SegFormer \cite{xie2021segformer}. 
\begin{equation}
\begin{aligned}
  & F_{i,j} = Upsample(Linear_i(C_i,C)(F_{i,j})), \forall{i, j}\\
  & \hat{F}_j = Linear(4C,C)(Concat(F_{i,j})), \forall{i}
  \label{eq:5}
\end{aligned}
\end{equation}
where $Linear(C_{in},C_{out})(\cdot)$ refers to a linear layer with $C_{in}$ and $C_{out}$ as input and output dimensions respectively, and $Upsample(\cdot)$ refers to upsampling features to 1/4th.

Here we obtain the feature maps $\hat{F}_{0}$ and $\hat{F}_{1}$ from each bi-temporal image. And then $\hat{F}_{0}$ and $\hat{F}_{1}$ are aggregated and projected to the ultimate change map $M\in \mathbb{R}^{2 \times H \times W}$, which can be formulated as:
\begin{equation}
\begin{aligned}
  & \hat{F} = Fuse(\hat{F}_{0}, \hat{F}_{1})\\
  & Y = Project(\hat{F})
  \label{eq:6}
\end{aligned}
\end{equation}
where $Fuse(\cdot)$ and $Project(\cdot)$ refer to the fusion layer and projection layer, respectively. In particular, the projection layer consists of two convolutional layers.

\subsection{ChangerVanilla}
For better comparisons, we first build a baseline model, ChangerVanilla. ChangerVanilla has no interaction layer and uses a simple concat operation as the fusion layer. Let $x$ denotes the feature map, we formulate the interaction and fusion layer of ChangerVanilla as:
\begin{equation}
\begin{aligned}
  & InterAct_{vanilla}(x_i) = Identity(x_i)\\
  & Fuse_{vanilla}(x_0, x_1) = Concat([x_0, x_1])
  \label{eq:7}
\end{aligned}
\end{equation}

Then, for the specific exploration of feature interaction layers, there are many complex strategies can be adapted. Here, however, we want to demonstrate that using only simple modules, even parameter-free interaction operations, can effectively improve the performance of change detection models, which has rarely been discussed in previous related studies. Specifically, we throw in two embarrassingly simple interaction methods: AD and feature ``exchange''.

\subsection{ChangerAD}
We abstract the aggregation-distribution style feature interaction from co-attention and some similar mechanisms \cite{lu2016hierarchical, nguyen2018improved, chen2020bi, li2019selective}, as shown in \cref{fig:Components}(c). We refer to this variant as ChangerAD. The basic idea of AD is to project the bi-temporal features into a feature space and get the global co-feature, then, use the distributed attention maps adaptively re-weight each channel of the bi-temporal features.

Specifically, we first aggregate features from siamese branches via an element-wise summation. Then We use the global average pooling to generate global information. Furthermore, we take a MLP to extract the co-feature, and employ $sigmoid$ to obtain the ultimate two attention maps. The interaction layer of ChangerAD is formulated as:
\begin{equation}
\begin{aligned}
  & InterAct_{AD}(x_i) = x_i \cdot \sigma(\hat{x}_i) \\
  & \hat{x} = MLP(C_i,2C_i)(GAP(x_0 + x_1))
  \label{eq:8}
\end{aligned}
\end{equation}
where $x_0$ and $x_1$ refer to the bi-temporal features, and the $MLP(C_{in},C_{out})(\cdot)$ refers to a 2-layer MLP, with the first layer squeezing and the second layer expanding channel.

\begin{table*}
  \centering
  \begin{tabular}{@{}l|c|c|c|ccc|ccc@{}}
    \toprule[1pt]
    \multirow{2}{*}{Method} & \multirow{2}{*}{Backbone} & \multirow{2}{*}{\#Param (M)} & \multirow{2}{*}{MACs (G)} & \multicolumn{3}{c|}{S2Looking} & \multicolumn{3}{c}{LEVIR-CD}\\
    & & & & Precision & Recall & F1 & Precision & Recall & F1\\
    \midrule
    FC-EF \cite{daudt2018fully} & - & 1.35 & 12.48 & 81.36 & 8.95 & 7.65 & 86.91 & 80.17 & 83.40\\
    FC-Siam-Conc \cite{daudt2018fully} & - & 1.54 & 19.47 & 68.27 & 18.52 & 13.54 & 91.99 & 76.77 & 83.69\\
    FC-Siam-Diff \cite{daudt2018fully} & - & 1.35 & 17.06 & 83.29 & 15.76 & 13.19 & 89.53 & 83.31 & 86.31\\
    DTCDSCN \cite{liu2020building} & SE-Res34 & 41.07 & 60.87 & 68.58 & 49.16 & 57.27 & 88.53 & 86.83 & 87.67\\
    STANet-Base \cite{chen2020spatial} & ResNet18 & - & - & 25.75 & 56.29 & 35.34 & 79.20 & 89.10 & 83.90\\
    STANet-BAM \cite{chen2020spatial} & ResNet18 & 12.18 & 49.16 & 31.19 & 52.91 & 39.24 & 81.50 & 90.40 & 85.70\\
    STANet-PAM \cite{chen2020spatial} & ResNet18 & 12.21 & 50.21 & 38.75 & 56.49 & 45.97 & 83.81 & 91.00 & 87.26\\
    CDNet \cite{chen2021adversarial}& ResNet18 & 14.33 & - & 67.48 & 54.93 & 60.56 & 91.60 & 86.50 & 89.00\\
    BiT \cite{chen2021efficient} & ResNet18 & 3.55 & 33.89 & 72.64 & 53.85 & 61.85 & 89.24 & 89.37 & 89.31\\
    ChangeFormer* \cite{bandara2022transformerbased}& MiT-b1 & 20.75 & 22.70 & 72.82 & 56.13 & 63.39 & 92.59 & 89.68 & 91.11\\
    \hline
    \hline
    ChangerVanilla & ResNet18 & 11.39 & 23.65 & \textbf{72.59} & 58.25 & 64.63 & 92.66 & \textbf{89.60} & 91.10\\
    ChangerAlign & ResNet18 & 11.39 & 23.71 & 71.62 & {\color{blue} 60.06} & \textbf{65.33} & {\color{blue} 93.30} & 89.59 & \textbf{91.41}\\
    ChangerAD & ResNet18 & 11.46 & 23.71 & {\color{red} 74.21} & \textbf{58.97} & {\color{blue} 65.72} & {\color{red} 93.34} & {\color{blue} 90.12} & {\color{blue} 91.70}\\
    ChangerEx & ResNet18 & 11.39 & 23.71 & {\color{blue} 73.59} & {\color{red} 60.15} & {\color{red} 66.20} & \textbf{92.97} & {\color{red} 90.61} & {\color{red} 91.77}\\
    \bottomrule[1pt]
  \end{tabular}
  \caption{Comparisons of Changer with other change detection method on parameters, computational cost(MACs), precision(\%), recall(\%) and F1-Score(\%) on S2Looking and LEVIR-CD. The MACs are computed  with RGB input image at the resolution of 512×512. The symbol ``*'' means our re-implemented results. Color convention: {\color{red} best}, {\color{blue} 2nd-best}, and \textbf{3rd-best} for Changer models.}
  \label{table1}
\end{table*}

\begin{figure*}
  \centering
  \begin{subfigure}{0.138\linewidth}
    \includegraphics[width=1.0\linewidth]{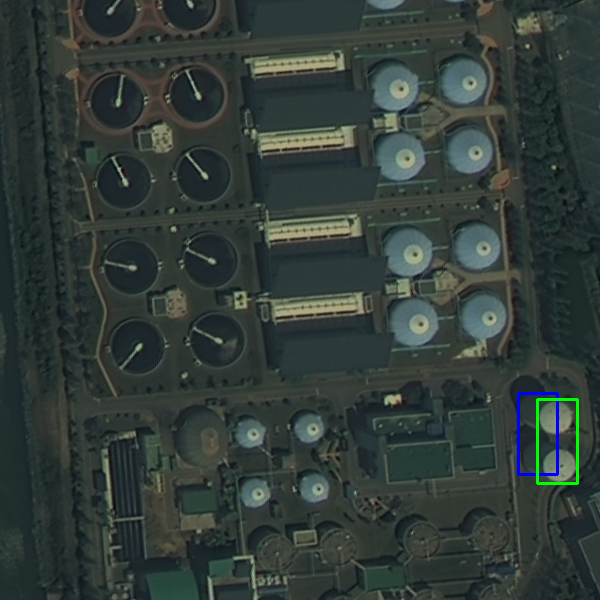}
    \label{fig:short-a1}
  \end{subfigure}
  \hfill
  \begin{subfigure}{0.138\linewidth}
    \includegraphics[width=1.0\linewidth]{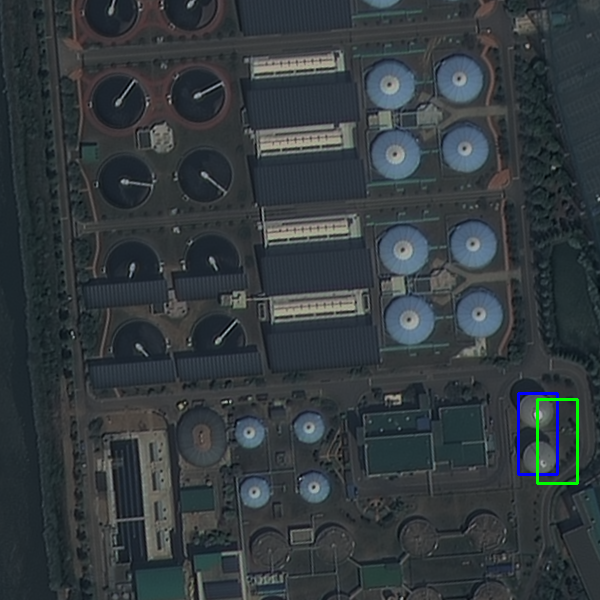}
    \label{fig:short-a2}
  \end{subfigure}
  \hfill
  \begin{subfigure}{0.138\linewidth}
    \includegraphics[width=1.0\linewidth]{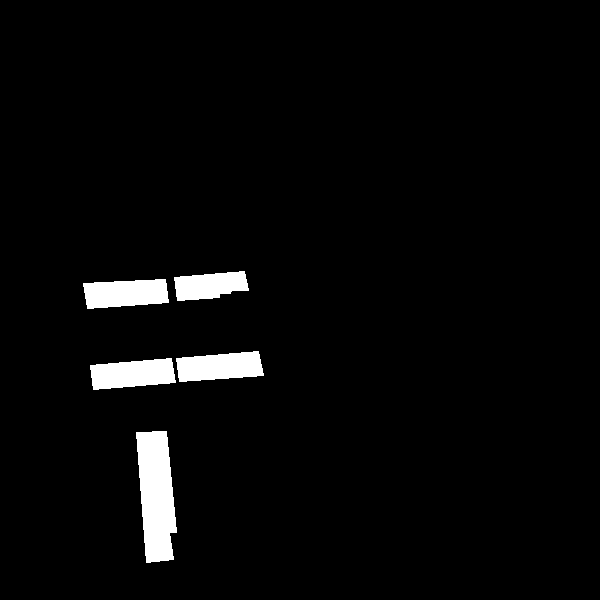}
    \label{fig:short-a3}
  \end{subfigure}
  \hfill
  \begin{subfigure}{0.138\linewidth}
    \includegraphics[width=1.0\linewidth]{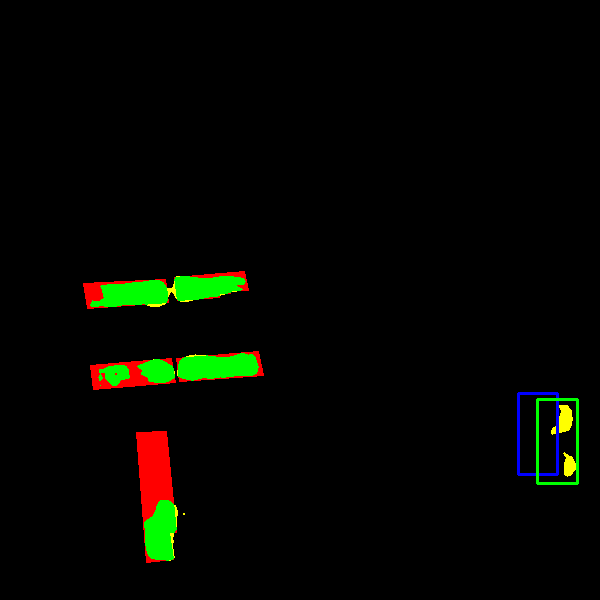}
    \label{fig:short-a4}
  \end{subfigure}
  \hfill
  \begin{subfigure}{0.138\linewidth}
    \includegraphics[width=1.0\linewidth]{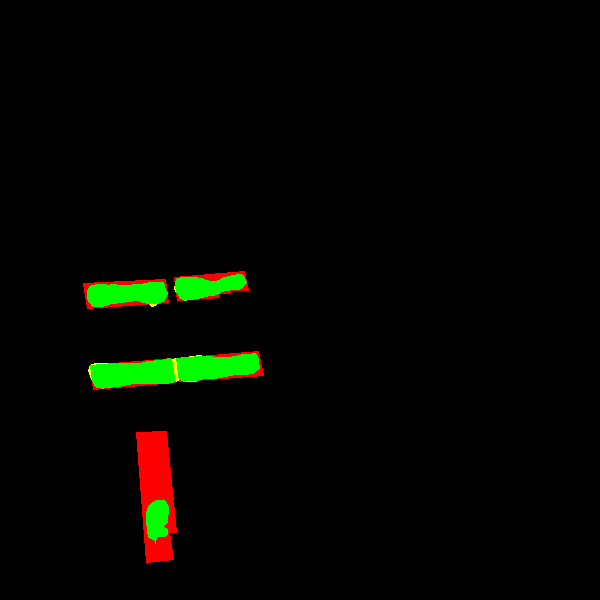}
    \label{fig:short-a5}
  \end{subfigure}
  \hfill
  \begin{subfigure}{0.138\linewidth}
    \includegraphics[width=1.0\linewidth]{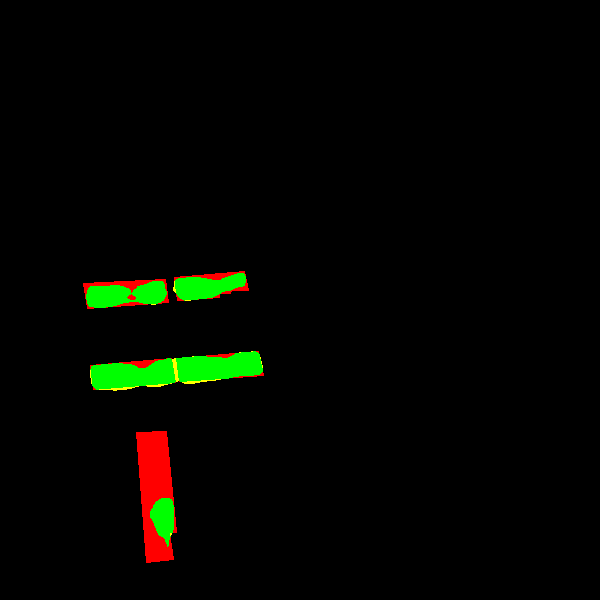}
    \label{fig:short-a6}
  \end{subfigure}
  \hfill
  \begin{subfigure}{0.138\linewidth}
    \includegraphics[width=1.0\linewidth]{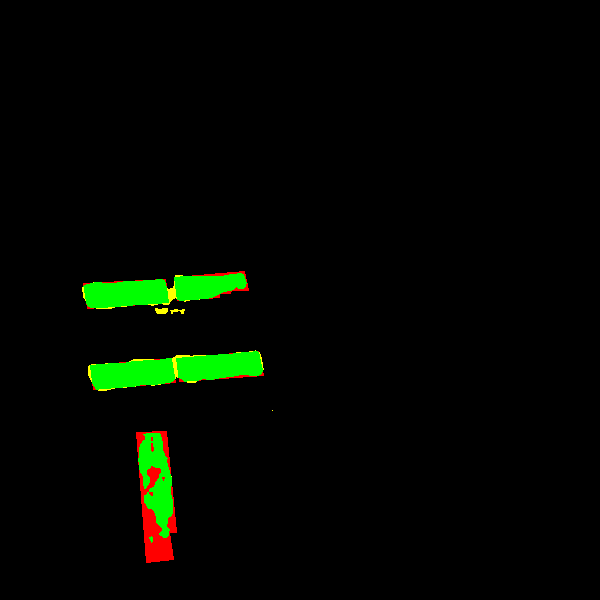}
    \label{fig:short-a7}
  \end{subfigure}
  
  \hfill
  \begin{subfigure}{0.138\linewidth}
    \includegraphics[width=1.0\linewidth]{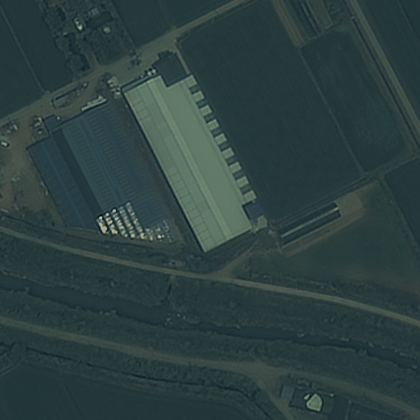}
    \caption{$t_1$ Image}
    \label{fig:short-b1}
  \end{subfigure}
  \hfill
  \begin{subfigure}{0.138\linewidth}
    \includegraphics[width=1.0\linewidth]{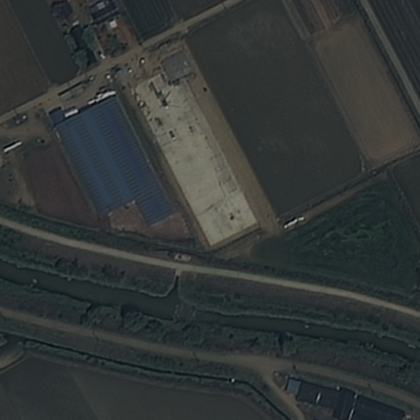}
    \caption{$t_2$ Image}
    \label{fig:short-b2}
  \end{subfigure}
  \hfill
  \begin{subfigure}{0.138\linewidth}
    \includegraphics[width=1.0\linewidth]{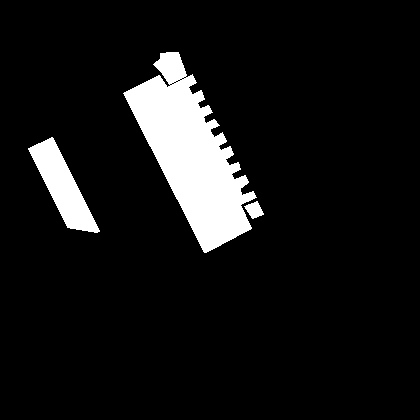}
    \caption{Ground Truth}
    \label{fig:short-b3}
  \end{subfigure}
  \hfill
  \begin{subfigure}{0.138\linewidth}
    \includegraphics[width=1.0\linewidth]{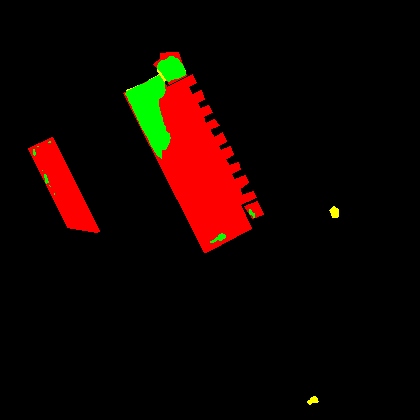}
    \caption{ChangerVanilla}
    \label{fig:short-b4}
  \end{subfigure}
  \hfill
  \begin{subfigure}{0.138\linewidth}
    \includegraphics[width=1.0\linewidth]{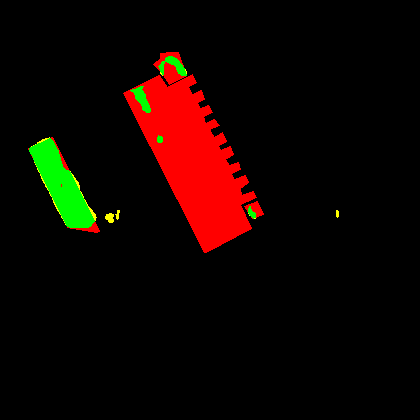}
    \caption{ChangerAlign}
    \label{fig:short-b5}
  \end{subfigure}
  \hfill
  \begin{subfigure}{0.138\linewidth}
    \includegraphics[width=1.0\linewidth]{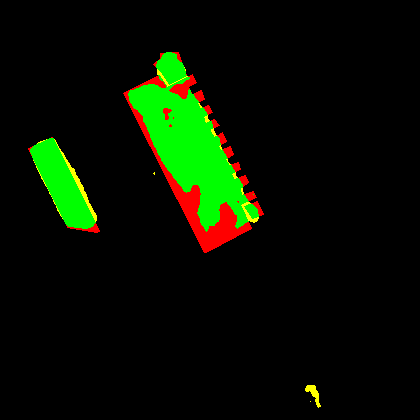}
    \caption{ChangerAD}
    \label{fig:short-b6}
  \end{subfigure}
  \hfill
  \begin{subfigure}{0.138\linewidth}
    \includegraphics[width=1.0\linewidth]{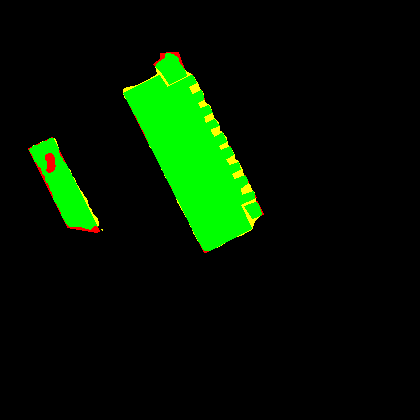}
    \caption{ChangerEx}
    \label{fig:short-b7}
  \end{subfigure}
  
  \caption{\textbf{Some visualization comparisons among Changer models on the S2Looking.} The rendered colors represent {\color{green}{true positives (TP)}}, {\color{yellow}false positives (FP)}, and {\color{red}false negatives (FN)}. Buildings with misregistration are framed out in the first row.}
  \label{fig:s2looking}
\end{figure*}


\begin{algorithm*}
\caption{Channel Exchange for ChangerEx, PyTorch-like Code}\label{alg:cap1}
\begin{algorithmic}
\State
\begin{lstlisting}[language=Python]
import torch
import torch.nn as nn

class ChannelExchange(nn.Module):
    def __init__(self, p=2):
        super().__init__()
        self.p = p

    def forward(self, x1, x2):
        N, C, H, W = x1.shape
        exchange_mask = torch.arange(C) % self.p == 0
        exchange_mask = exchange_mask.unsqueeze(0).expand((N, -1))
        out_x1, out_x2 = torch.zeros_like(x1), torch.zeros_like(x2)
        out_x1[~exchange_mask, ...] = x1[~exchange_mask, ...]
        out_x2[~exchange_mask, ...] = x2[~exchange_mask, ...]
        out_x1[exchange_mask, ...] = x2[exchange_mask, ...]
        out_x2[exchange_mask, ...] = x1[exchange_mask, ...]
        
        return out_x1, out_x2
\end{lstlisting}
\end{algorithmic}
\end{algorithm*}


\subsection{ChangerEx}
Feature ``exchange'' refers to: partial exchange between bi-temporal features, during feature extraction. Hence, the natural question is why exchange and why it is feasible.

\textbf{Why exchange?} On the one hand, contextual information of bi-temporal features can be perceived by mutual learning through feature exchange and the subsequent mix layers (\eg convolution or token mixer). On the other hand, through feature exchange and the subsequent layers, the distribution between the features of the two branches is more similar and automatic domain adaptation between the bi-temporal domains is achieved to some extent.

\textbf{Why is the exchange feasible?} While some studies have emphasised temporal information, we do not believe that bi-temporal change detection is strongly constrained by ``time''. The core of bi-temporal change learning is to train a change detector for images with the same spatial position but at different times. The temporal order is only to ensure that the appearance and disappearance of the target is interpretable, and in most definitions and applications in change detection, we are not concerned with whether it appears or disappears, but simply changes. In brief, the semantic correspondence constraint for bi-temporal images in the definition of change detection makes feature exchange feasible.

Then, an important issue is \textbf{where to exchange}. Because of the logical operations involved in determining whether or not to exchange certain features, we need to be cautious to avoid non-differentiable problem. For this, we considered two solutions. One is to convert the soft exchange mask to attention maps to guarantee the continuity of the gradient chain, and the other is to use an unlearnable way to exchange, i.e. predefine a hard exchange mask. In practice, we find that simple learnable exchange does not perform better than the parameter-free unlearnable exchange, and therefore we tend to use the latter. 

Let $x_0, x_1$ denote the bi-temporal features. $InterAct_{Ex}$ for $x_{0/1}$ can be formulated as:
\begin{equation}
x_{0/1}(n,c,h,w) =
\begin{cases}
  x_{0/1}(n,c,h,w),&M(n,c,h,w) = 0 \\
  x_{1/0}(n,c,h,w),&M(n,c,h,w) = 1
  \label{eq:9}
\end{cases}
\end{equation}
where $n$, $c$ and $hw$ index the batch, channel and space respectively. $M$ refers to the exchange mask consisting of 1 and 0, indicating exchange and non-exchange. 

Furthermore, we have tried feature exchange in two dimensions separately, the channel dimension and the spatial dimension, with the details given in what follows.

\subsubsection{Channel Exchange}
Channel exchange refers to exchanging features in channel dimension. Specifically, for \cref{eq:9}, the predefined $mask$ remains spatially consistent. After channel exchange, gradients are detached from the exchanged channel and back-propagated through the other ones.

\subsubsection{Spatial Exchange}
Relatively, spatial exchange refers to exchanging features in spatial dimension, and the $mask$ remains consistent on the channel. Considering that subsequent layers may be fully channel-wise MLP, a mix layer is optional.

The pseudo-code of channel exchange and spatial exchange are in \cref{alg:cap1} and Appendix \cref{alg:cap2}.
Finally, we combine these two feature exchange and propose ChangerEx, as shown in \cref{fig:ChangerEx}. In addition, feature exchange involves issues like how many features to exchange, at which stage to exchange, \etc which we will discuss later.

\subsection{Flow Dual-Alignment Fusion}
\label{section:FDAF}

Registration error is one of the most common challenges in change detection. Image registration is an essential part for change detection pre-processing. However, there are always more or less mis-alignment or side-looking problems, as shown in \cref{fig:s2looking}. Some work has tried implicit alignment, such as the use of global attention in \cite{chen2020spatial}. In this paper we use an explicit bi-directional alignment with optical flow. The FDAF introduces a task prior for change detection, i.e. discovering differences between images. Specifically, the bi-temporal feature maps are resampled through a deformable field to obtain their respective corrected features. Then we take the distance to the other original feature map separately and feed it into the subsequent forward propagation. In other words, the FDAF implements an explicit task transformation function where the object extraction task is converted into the change detection task. Mathematically, the features $x_0$, $x_1$ feed into $Fuse_{align}$ can be written as:
\begin{equation}
\begin{aligned}
  x = Concat([x_0(p+\Delta p_0) - x_1, x_1(p+\Delta p_1) - x_0])
  \label{eq:10}
\end{aligned}
\end{equation}
where $p$ enumerates the locations in $x_0$ or $x_1$, and the offset field $\Delta p$ is obtained by the $FlowNet$ (2-layer Conv):
\begin{equation}
\begin{aligned}
  \Delta p = FlowNet(Concat([x_0, x_1])
  \label{eq:11}
\end{aligned}
\end{equation}
The bi-linear interpolation \cite{jaderberg2015spatial} is used to compute the exact value of the features. We refer to MetaChanger with FDAF as ChangerAlign, as shown in \cref{fig:Components}(e).


\section{Experiments}

\subsection{Dataset}
\textbf{S2Looking.} S2Looking dataset \cite{shen2021s2looking} contains 5000 image pairs (1024 × 1024, 7:1:2 for train, eval and test) and more than 65,920 annotated instances of changes extracted from side-looking rural area satellite images, which were collected by optical satellites around the world. The images span 1-3 years with a resolution of 0.5-0.8 m/pixel.

\textbf{LEVIR-CD.} LEVIR-CD dataset \cite{chen2020spatial} contains 637 bitemporal RS image pairs and more than 31,333 annotated instances of changes, which were collected from Google Earth. Each image in the pairs is 1024 × 1024 with an image resolution of 0.5 m/pixels.

\subsection{Implementation detail}
We develop a change detection toolbox, Open-CD, based on PyTorch and open-mmlab related tools \cite{mmcv}. During training, we use the CE Loss and AdamW optimizer. The weight decay is set to 0.05 always. The $poly$ schedule with an initial learning rate of 0.001 is adopted. We use single Tesla V100 GPU for training and the batch size is set to 8. We train all Changer models for 80k and 40k iterations for S2Looking and LEVIR-CD dataset. For data augmentation, we use random crop, flip and photometric distortion. And we randomly exchange the order of the bi-temporal images. 

\subsection{Main results}
In \cref{table1}, We show the results of some SOTA and typical change detection methods, including ConvNets and Vision Transformers. The details of the compared methods are described in Appendix \ref{section:Appendix A}.
Our baseline ChangerVanilla achieves competitive performance than previous change detection methods, which demonstrates MetaChanger's effectiveness in overall architecture. The FDAF brings boost in F1-score, and its effect can also be observed in \cref{fig:s2looking}. ChangerAD and ChangerEx are built on top of ChangerAlign and make feature interactions during feature extraction. Compared to ChangerAlign, ChangerAD and ChangerEx achieved significant improvements with only slight or no increases in parameters and computational cost. Furthermore, the Changer models achieve more promising gains on the more challenging dataset S2Looking.


\begin{table}
  \centering
  \begin{tabular}{@{}cccc|c@{}}
    \toprule[1pt]
    \multicolumn{4}{c|}{Stages w/ Aggregation-Distribution} & \multirow{2}{*}{F1} \\
    Stage1 & Stage2 & Stage3 & Stage4\\
    \midrule
    & & & \checkmark & 65.63 \\
    & & \checkmark & \checkmark & {\color{blue} 65.69} \\
    & \checkmark & \checkmark & \checkmark & {\color{red} 65.72} \\
    \checkmark & \checkmark & \checkmark & \checkmark & 65.17 \\
    \bottomrule[1pt]
  \end{tabular}
  \caption{Ablation study on applying AD on different stages on S2Looking. \checkmark means that AD is used at this stage.}
  \label{table_AD}
\end{table}

\begin{table}
  \centering
  \begin{tabular}{@{}cccc|c@{}}
    \toprule[1pt]
    \multicolumn{4}{c|}{Stages w/ Channel Exchange} & \multirow{2}{*}{F1} \\
    Stage1 & Stage2 & Stage3 & Stage4\\
    \midrule
    & & & \checkmark & {\color{blue} 65.71} \\
    & & \checkmark & \checkmark & {\color{red} 65.91} \\
    & \checkmark & \checkmark & \checkmark & 65.53 \\
    \checkmark & \checkmark & \checkmark & \checkmark & 65.22 \\
    \bottomrule[1pt]
  \end{tabular}
  \caption{Ablation study on applying channel exchange on different stages on S2Looking. \checkmark means that 1/2 channel is exchanged at this stage.}
  \label{table2}
\end{table}

\begin{table}
  \centering
  \begin{tabular}{@{}cccc|c@{}}
    \toprule[1pt]
    \multicolumn{4}{c|}{Stages w/ Spatial Exchange} & \multirow{2}{*}{F1}\\
    Stage1 & Stage2 & Stage3 & Stage4\\
    \midrule
    & & & \checkmark & 65.38 \\
    & & \checkmark & \checkmark &  65.60\\
    & \checkmark & \checkmark & \checkmark & {\color{red} 66.11} \\
    \checkmark & \checkmark & \checkmark & \checkmark & {\color{blue} 65.66} \\
    \bottomrule[1pt]
  \end{tabular}
  \caption{Ablation study on applying spatial exchange on different stages on S2Looking. \checkmark means that 1/2 spatial embedding is exchanged at this stage.}
  \label{table3}
\end{table}


\subsection{Ablation Studies}
To delve into MetaChanger and its variants, especially ChangerEx, we conducted comprehensive experiments for the following questions. If not specified, $ResNet18\_V1c$ (without pretrained) is used as the backbone network.

\textbf{Which stage to exchange.} We insert interaction layers including AD and channel/spacial exchange at different stages. As shown in \cref{table_AD}, Inserting AD layer in the last three stages achieves the best performance, while inserting the interaction layer in the first stage hurts the performance. A similar situation occurs in channel exchange, as shown in \cref{table2}: the best results are obtained with the interaction layer in the latter two stages only. In spatial exchange, however, the situation is different, with the best two settings interacting in the earlier stages, as listed in \cref{table3}. An intuition is that in the shallow layers of the network, there is a high spatial resolution and a lower channel dimension, which is more suitable for spatial interactions; and vice versa. Based on this observation, we use spatial exchange in the shallow layers and channel exchange in the deeper layers for ChangerEx, as shown in \cref{fig:ChangerEx}(a).

\textbf{What ratio of features should be exchanged.}
As shown in \cref{table5}, we try various exchange ratios, from 1/32 to 1/2. We find that the difference in performance of ChangerEx at different exchange ratios is relatively slight. Thus, in feature exchange, the ratio of features exchanged is not the determining factor; the emphasis is on the presence or absence of exchange. This ablation also suggests that feature exchange is insensitive, even when some additional hyper-parameters are introduced.

\textbf{How to choose the size of the exchange window in spatial exchange.} A worthwhile consideration is whether small exchange patch in spatial exchange would disrupt the original spatial structure. We therefore tried using different window sizes in spatial exchange, from 1 × 1 to 8 × 8, as listed in the \cref{table7}. We find that the window size is a robust hyper-parameter and that a more complete original spatial structure does not result in a performance gain.

\textbf{Learnable exchange.} We try the learnable exchange method in the channel and spatial dimensions respectively. In our implementation, we generate an exchange map based on the distance between the soft attention maps of the two branches, with the smaller half being exchanged and the larger half retained. And the two attention maps also need to be exchanged if their corresponding features are exchanged. As \cref{table6} lists, the learnable channel exchange shows a slight improvement over the unlearnable one, but the performance of the learnable spatial exchange drops dramatically.

We have only tested simple learnable exchanges here, and this part deserves further exploration in future work.

\textbf{MetaChanger with more complex backbones.} To further illustrate the generalizability of ChangerEx to different network architectures, we replaced the backbone with the more complex networks, ResNeSt50 and ResNeSt101. As listed in \cref{table8}, the networks with feature exchange outperform all the baselines significantly, demonstrating that the ChangerEx can generalize well on various models, especially in more challenging dataset.

We find that ChangerEx can deliver higher gains to complex models in large-scale datasets, which is promising. Specifically. ChangerEx with ResNeSt101 leads to a performance gain of 1.68 (achieving a F1-Score of 67.61).

\begin{figure}
  \centering
  \begin{subfigure}{0.322\linewidth}
    \includegraphics[width=1.0\linewidth]{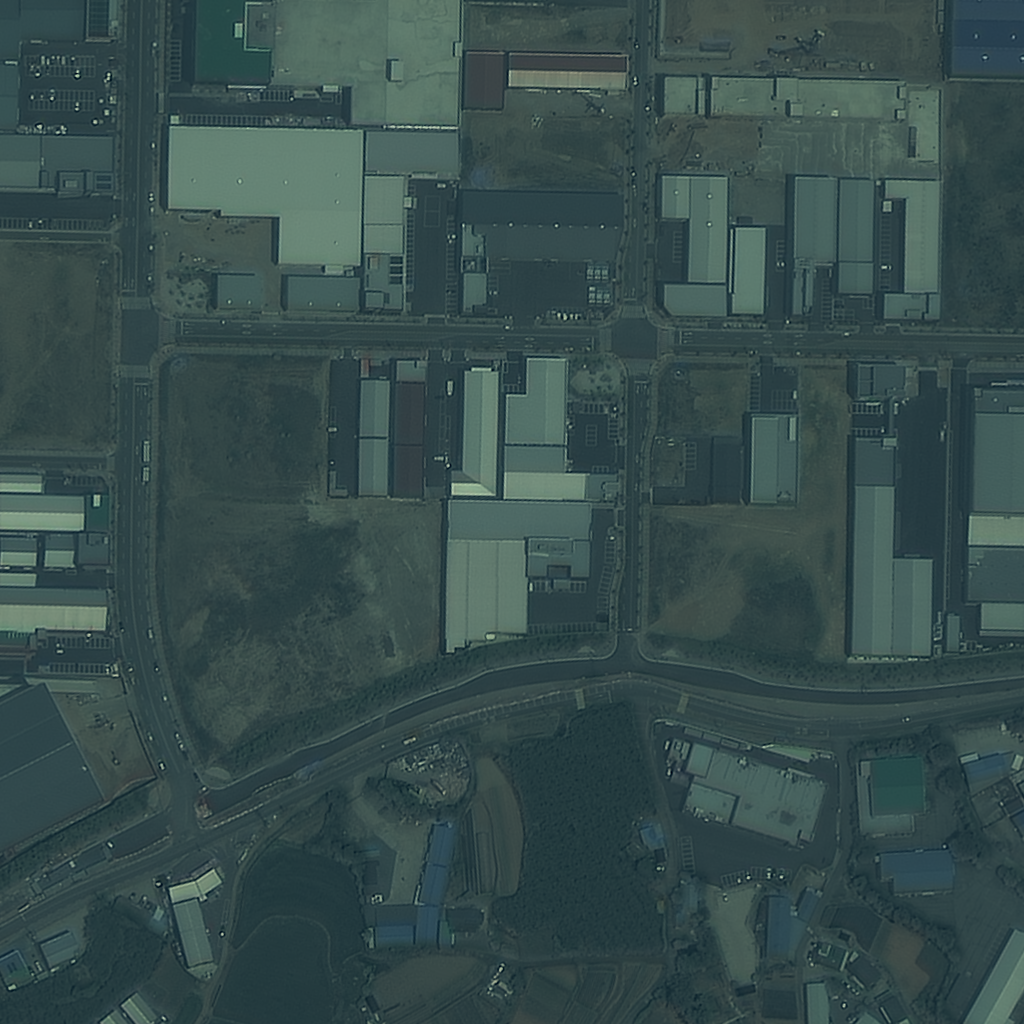}
    \label{fig:cam-a1}
  \end{subfigure}
  \hfill
  \begin{subfigure}{0.322\linewidth}
    \includegraphics[width=1.0\linewidth]{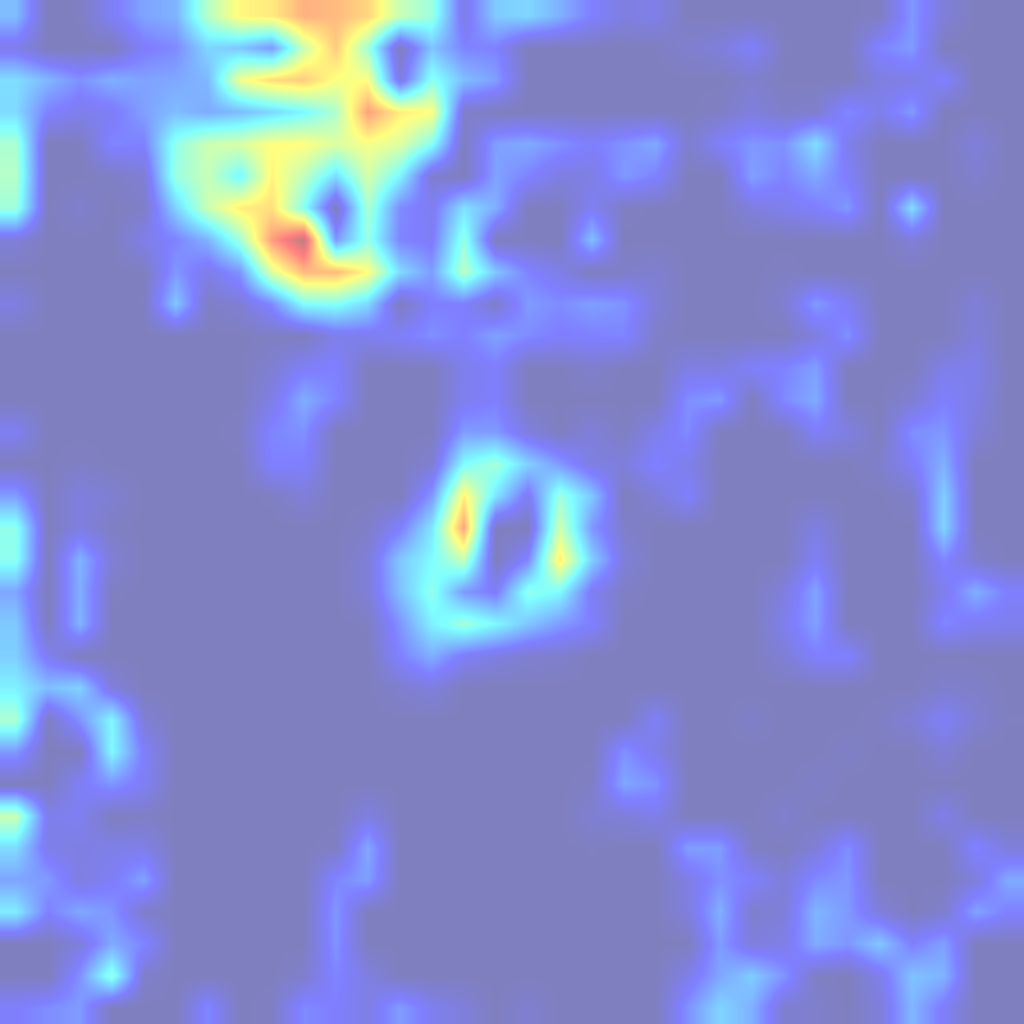}
    \label{fig:cam-a2}
  \end{subfigure}
  \hfill
  \begin{subfigure}{0.322\linewidth}
    \includegraphics[width=1.0\linewidth]{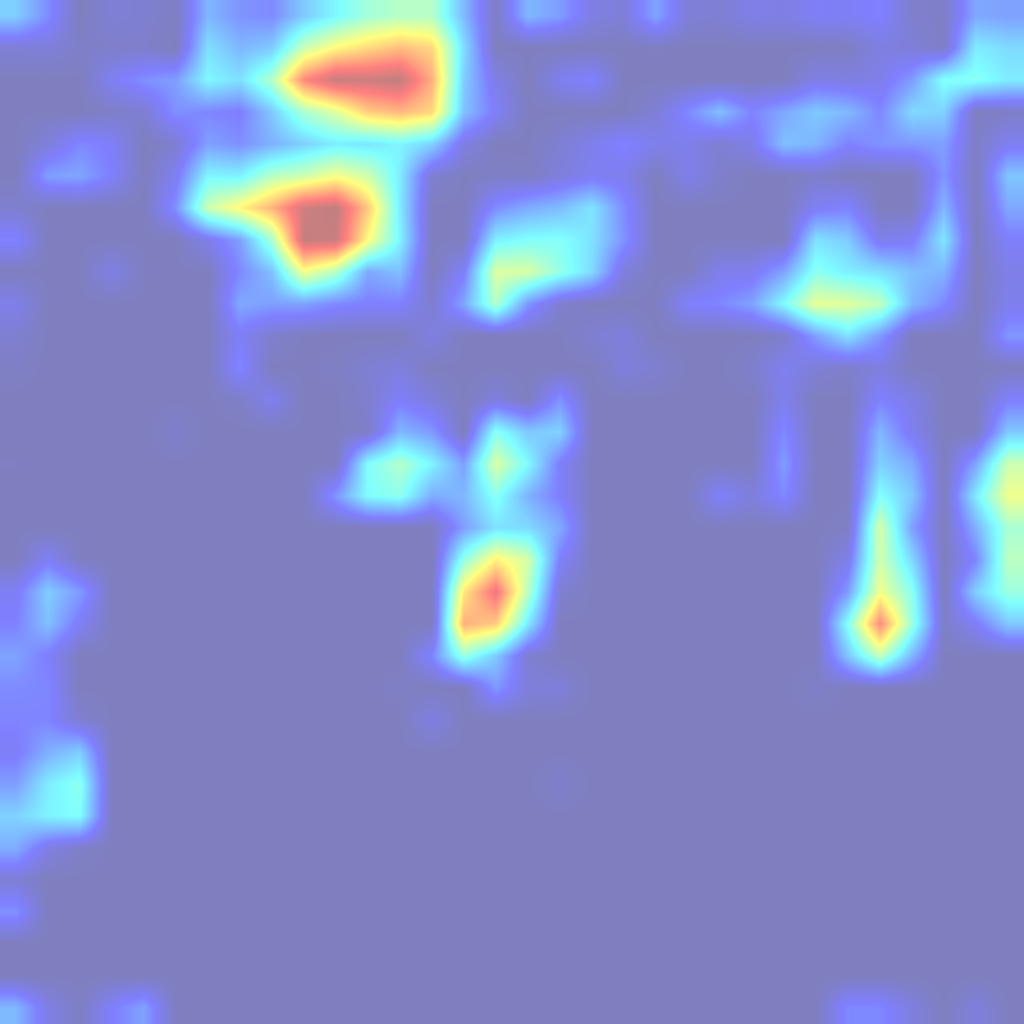}
    \label{fig:cam-a3}
  \end{subfigure}
  \hfill
  \begin{subfigure}{0.322\linewidth}
    \includegraphics[width=1.0\linewidth]{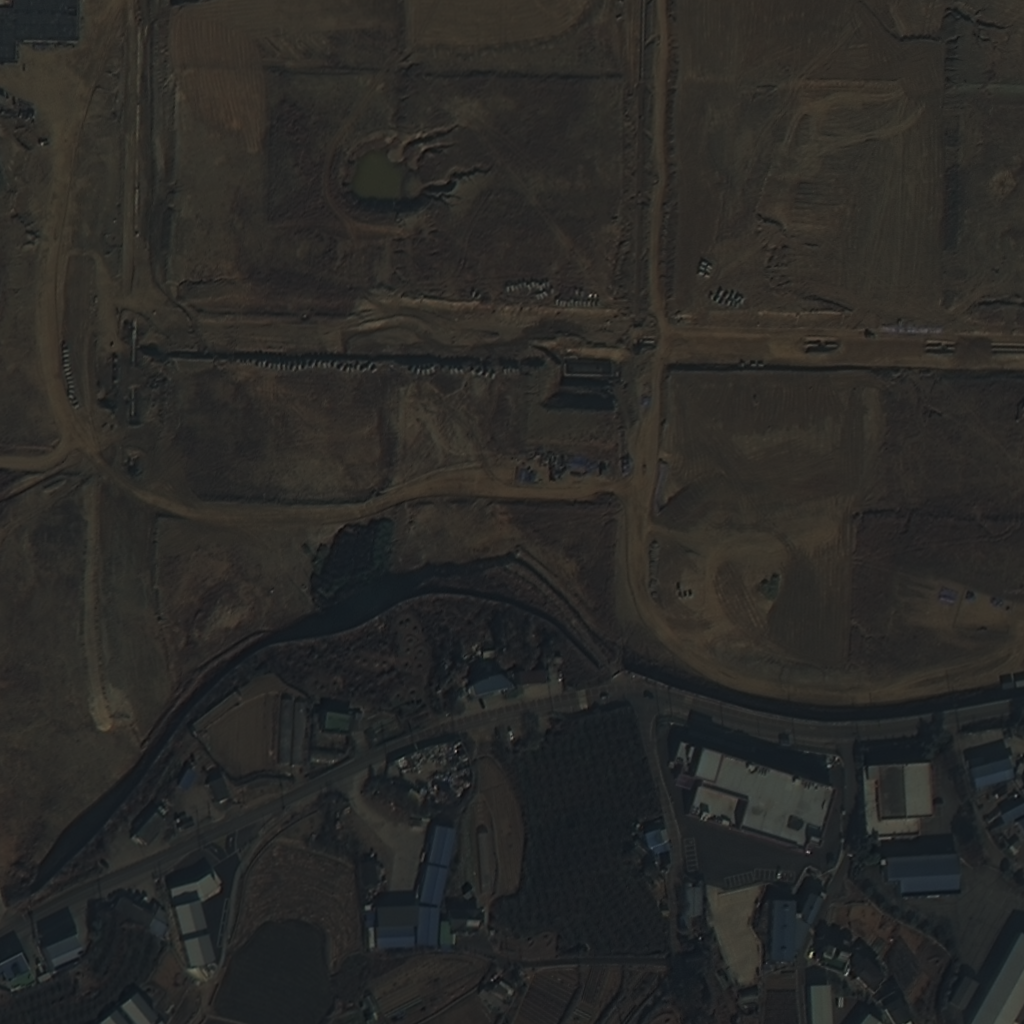}
    \caption{$t_1/t_2$ Image}
    \label{fig:cam-a4}
  \end{subfigure}
  \hfill
  \begin{subfigure}{0.322\linewidth}
    \includegraphics[width=1.0\linewidth]{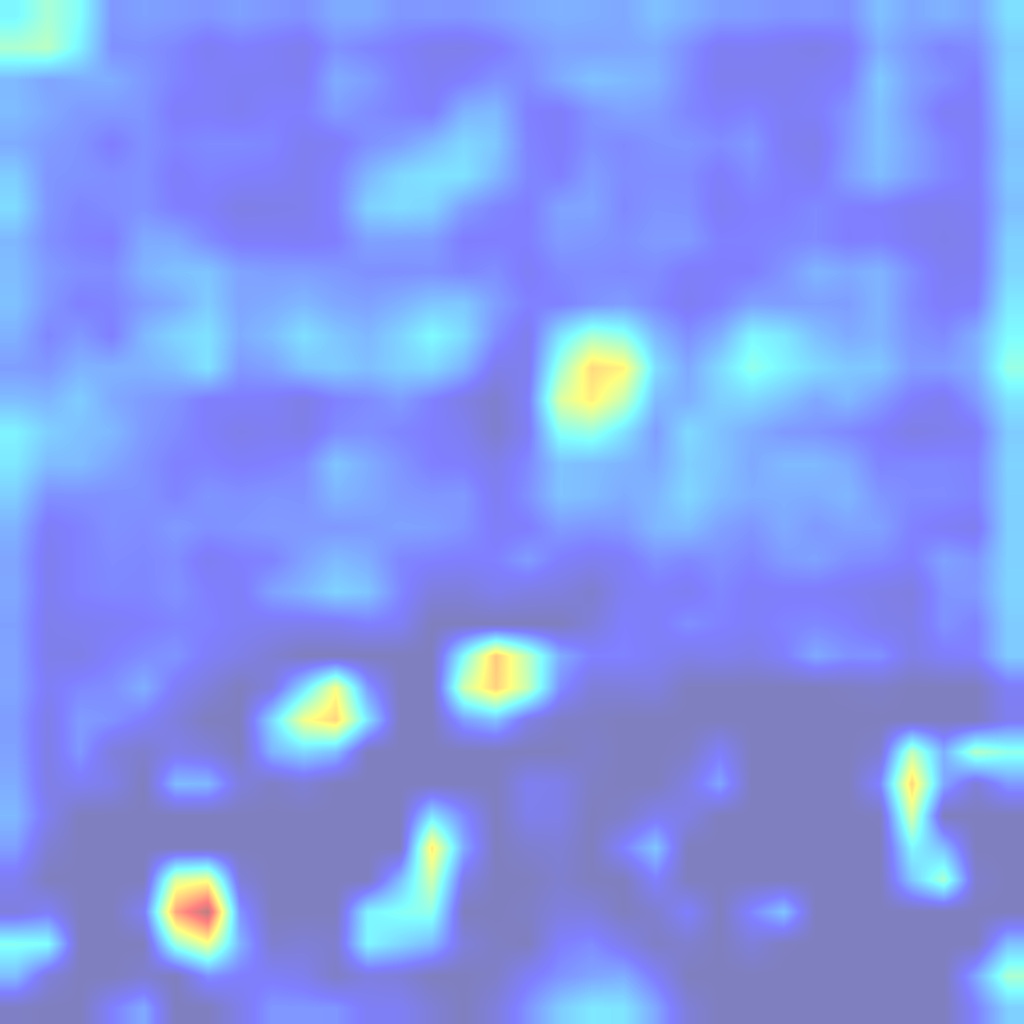}
    \caption{w/o exchange}
    \label{fig:cam-a5}
  \end{subfigure}
  \hfill
  \begin{subfigure}{0.322\linewidth}
    \includegraphics[width=1.0\linewidth]{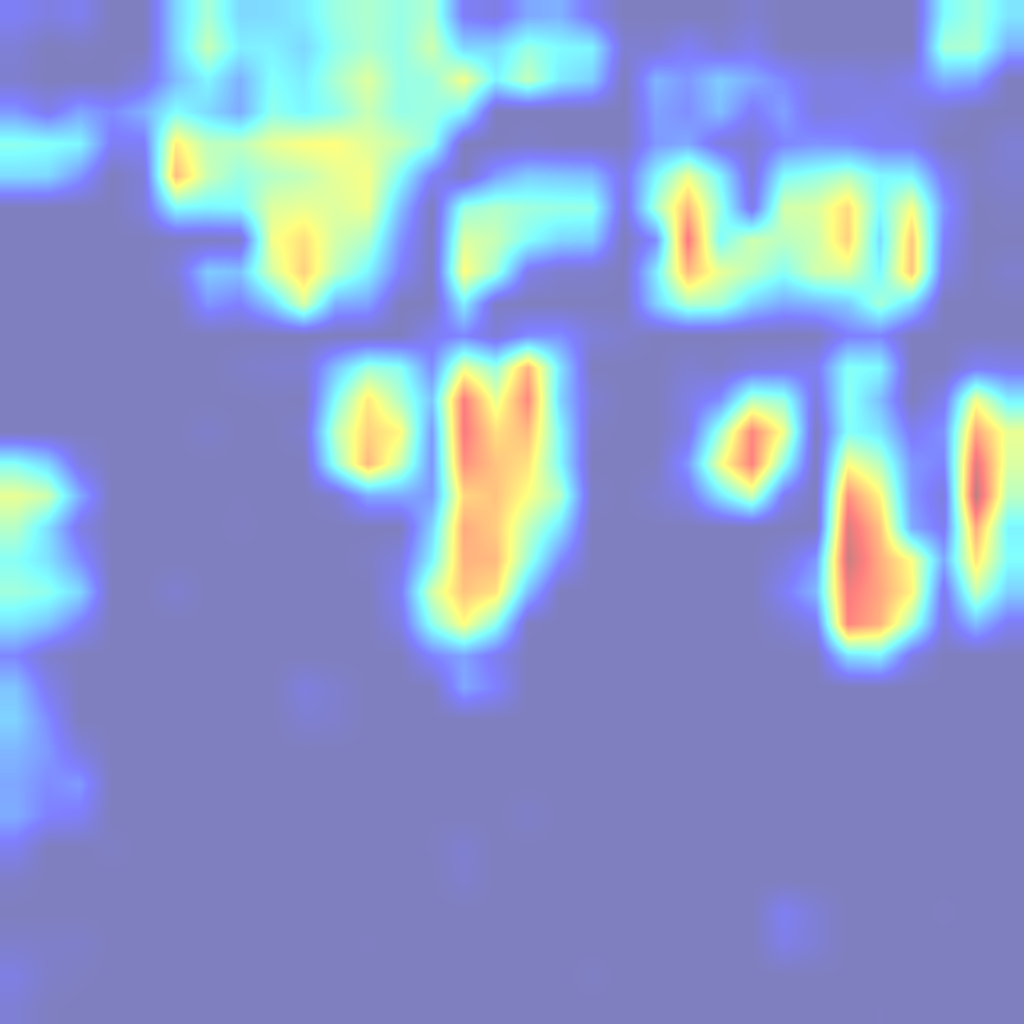}
    \caption{w/ exchange}
    \label{fig:cam-a6}
  \end{subfigure}
  
  \caption{\textbf{Grad-CAM visualization results (for all of the pixels from ``change'' category).} We compare the visualization results of ChangerAlign and ChangerEx (both with ResNeSt101). The grad-CAM visualization is calculated for the last stage outputs.}
  \label{fig:cam}
\end{figure}

\textbf{Why exchange work?} To further explore why ChangerEx is effective, we visualize the Changer models with and without ``exchange'' separately, using grad-CAM \cite{selvaraju2017grad}. As shown in \cref{fig:cam}, most of the buildings in the upper half of the image disappear from $t_1$ to $t_2$. In the $t_1$-heat map with no exchange, only very few building areas are activated. In the $t_1$-heat map with exchange, the situation improves but is similar overall. However, an interesting phenomenon is that the areas with buildings in $t_1$ are activated in the $t_2$-heat map with feature exchange. In other words, the perceptual targets that are lost in $t_1$ are reactivated in $t_2$.

Another possible explanation is that feature exchange increases the diversity of the samples, achieving a kind of intra-network data augmentation. The order of appearance and disappearance of targets of interest in the bi-temporal features is shuffled, but with still keeping the strict semantic maintenance and semantic correspondence.


\begin{table}
  \centering
  \begin{tabular}{@{}c|ccc@{}}
    \toprule[1pt]
    Exchange Ratio & Precision & Recall & F1 \\
    \midrule
    1/32 & 72.31 & \textbf{60.72} & 66.01 \\
    1/16 & 72.68 & 60.50 & 66.03 \\
    1/8 & 72.81 & 60.43 & 66.05 \\
    1/4 & 72.58 & 60.40 & 66.04 \\
    1/2 & \textbf{73.59} & 60.15 & \textbf{66.20} \\
    \bottomrule[1pt]
  \end{tabular}
  \caption{Ablation study on exchange ratios used in spatial and channel exchange on S2Looking.}
  \label{table5}
\end{table}

\begin{table}
  \centering
  \begin{tabular}{@{}c|ccc@{}}
    \toprule[1pt]
    Window Size & Precision & Recall & F1 \\
    \midrule
    8 × 8 & 72.83 & 60.38 & 66.02 \\
    4 × 4 & 72.75 & \textbf{60.41} & 66.01  \\
    2 × 2 & 72.98 & 60.40 & 66.10 \\
    1 × 1 & \textbf{73.59} & 60.15 & \textbf{66.20} \\
    \bottomrule[1pt]
  \end{tabular}
  \caption{Ablation study on different window size options for spatial exchange on S2Looking.}
  \label{table7}
\end{table}

\begin{table}
  \centering
  \begin{tabular}{@{}cc|ccc@{}}
    \toprule[1pt]
    Exchange & Learnable & Precision & Recall & F1 \\
    \midrule
    \multirow{2}{*}{Channel} & \checkmark & 73.77 & \textbf{59.61} & \textbf{65.94} \\
    & \ding{55} & \textbf{74.19} & 59.29 & 65.91 \\
    \hline
    \multirow{2}{*}{Spatial} & \checkmark & 73.41 & 56.53 & 63.88\\
    & \ding{55} & \textbf{74.51} & \textbf{59.41} & \textbf{66.11}\\
    \bottomrule[1pt]
  \end{tabular}
  \caption{Ablation study on whether use learnable exchange for spatial/channel exchange on S2Looking.}
  \label{table6}
\end{table}

\begin{table}
  \centering
  \begin{tabular}{@{}llccc@{}}
    \toprule[1pt]
    Backbone & Setting & S2Looking & LEVIR-CD \\
    \midrule
    \multirow{3}{*}{ResNet18} & Vanilla & 64.63 & 91.10\\
    & +FDAF & 65.33 & 91.41\\
    & +FDAF +Ex & \textbf{66.20} & \textbf{91.77}\\
    \hline
    \hline
    \multirow{2}{*}{ResNeSt50} & Vanilla & 65.04 & 91.97 \\
    & +FDAF & 65.44 & 92.04\\
    & +FDAF +Ex & \textbf{67.18} & \textbf{92.19}\\
    \hline
    \hline
    \multirow{2}{*}{ResNeSt101} & Vanilla & 65.66 & 91.99 \\
    & +FDAF & 65.93 & 92.07\\
    & +FDAF +Ex & \textbf{67.61} & \textbf{92.33}\\
    \bottomrule[1pt]
  \end{tabular}
  \caption{F1-Score on S2Looking and LEVIR-CD of various CNN backbone with Changer. ``Ex'' means ``exchange''.}
  \label{table8}
\end{table}

\section{Conclusion}
In this paper we propose MetaChanger to explore the effect of feature interactions in change detection. To verify the effectiveness of feature interaction, we deliberately specify interaction layer as extremely simple aggregation-distribution and feature ``exchange'' for MetaChanger.
It is found that the derived ChangerAD and ChangerEx can achieve competitive performance on multiple change detection datasets. Extensive ablation studies demonstrate the robustness and extensibility of ChangerEx. 

In the future, we will further evaluate MetaChanger under more different learning settings and related tasks, such as semantic change detection. We hope this work can inspire more future research devoted to improving the MetaChanger especially the interaction methods.

{\small
\bibliographystyle{ieee_fullname}
\bibliography{egbib}
}


\setcounter{algorithm}{0}
\renewcommand{\thealgorithm}{B\arabic{algorithm}}



        

\begin{algorithm*}
\caption{Spatial Exchange for ChangerEx, PyTorch-like Code}\label{alg:cap2}
\begin{algorithmic}
\State
\begin{lstlisting}[language=Python]
import torch
import torch.nn as nn

class SpatialExchange(nn.Module):
    def __init__(self, p=2):
        super().__init__()
        self.p = p

    def forward(self, x1, x2):
        N, C, H, W = x1.shape
        exchange_mask = torch.arange(w) % p == 0
        out_x1, out_x2 = torch.zeros_like(x1), torch.zeros_like(x2)
        out_x1[..., ~exchange_mask] = x1[..., ~exchange_mask]
        out_x2[..., ~exchange_mask] = x2[..., ~exchange_mask]
        out_x1[..., exchange_mask] = x2[..., exchange_mask]
        out_x2[..., exchange_mask] = x1[..., exchange_mask]

        return out_x1, out_x2
\end{lstlisting}
\end{algorithmic}
\end{algorithm*}

\newpage
\appendix

\section{Compared methods}
\label{section:Appendix A}

\textbf{FC-EF, FC-Siam-Conc and FC-Siam-Diff} \cite{daudt2018fully} are three classification-based UNet-like models. FC-EF uses early fusion to directly concatenate bi-temporal images, FC-Siam-Conc and FC-Siam-Diff use siamese encoders and use concatenation and difference to fuse features respectively.

\textbf{DTCDSCN} \cite{liu2020building} is a multi-task model, which can accomplish both change detection and semantic segmentation at the same time. It also introduces a dual-attention module to exploit the interdependencies between channels and spatial positions, improving the feature representation.

\textbf{STANet} \cite{chen2020spatial} is a siamese network with spatial-temporal attention designed to explore spatial-temporal relationships for change detection. It includes a base model (\textbf{STANet-Base}) that uses a weight-sharing CNN feature extractor, and optimise models through metric method. \textbf{STANet-BAM} and \textbf{STANet-PAM} equip the basic spatial-temporal attention module (like self-attention) and the pyramid spatial-temporal attention module on top of STANet-Base

\textbf{CDNet} \cite{chen2021adversarial} is a well-tuned siamese CNN model. CDNet is used with an instance-level data augmentation, which can generate bi-temporal images that contain changes involving plenty and diverse synthesized building instances by leveraging generative adversarial training.

\textbf{BiT} \cite{chen2021efficient} is a hybrid model of CNN and transformer. It uses the convolutional blocks at the shallow layers and the transformer blocks (with cross-attention) at the deeper layers, which can effectively model contexts within the spatial-temporal domain.

\textbf{ChangeFormer} \cite{chen2021efficient} is a transformer-based siamese network. ChangeFormer combines a hierarchical transformer encoder with a MLP decoder in a siamese network to effectively render long-range details.

\section{Pseudo code}
The pseudo-code for spatial exchange is shown in \cref{alg:cap2}. The exchange mask is obtained in the $W$ dimension and broadcast to the $C \times H \times W$, which allows more stable testing at different scales than obtaining it in the $HW$ dimension.

\end{document}